# Evaluation and Comparison of Deep Learning Methods for Pavement Crack Identification with Visual Images

Kai-Liang LU[‾]

*Jiangsu Automation Research Institute Shanghai Branch, Shengxia Road NO.666, Pudong, Shanghai 201210, P.R. China*
*lukailiang@163.com*

Pavement crack identification with visual images via deep learning algorithms, compared to the contact detection techniques, has the advantages of not being limited by the material of the tested object, fast speed and low cost. The fundamental frameworks and typical model architectures of transfer learning (TL), encoder-decoder (ED), generative adversarial networks (GAN) as well as their common modules were firstly reviewed, then the evolution of CNN backbone models and GAN models were summarized. The crack classification, segmentation performance and effect were tested on SDNET2018 or CFD public data sets. In the aspect of patch sample classification, the fine-tuned TL models can be equivalent to or even slightly better than the ED models in accuracy, and the predicting time is faster; In the aspect of accurate crack location, both ED and GAN algorithms can achieve pixel-level segmentation and is expected to be detected in real time on low computing power platform. A quantitative evaluation indicator is advised for crack ground truth (GT) semi-accurate labeling, when per-pixel accurate labeling is difficult or infeasible. Furthermore, a weakly supervised learning framework of combined TL-SSGAN and its performance enhancement measures are proposed, which can maintain comparable crack identification performance with the supervised learning, while greatly reducing the number of labeled samples needed.

## 1. Introduction

Structural cracks identification or detection with visual images[1-3] is non-contact, not restricted by the material of the tested objects and can be automatically detected online end-to-end. Compared to non-destructive testing (NDT)[4-5] or structural health monitoring (SHM)[6] techniques, it has obvious advantages of fast speed, low cost and high identification accuracy. Therefore, it has a wide application prospect in routine inspection and other preventive detection or monitoring scenarios, where a defective object whose geometrical topological form can be abstracted into a 2D plane form.

The image-based surface crack identification technology via hand-crafted feature engineering, machine learning or deep learning methods, can realize the qualitative and quantitative detection of cracks: 1) Classification, to judge whether there is a crack or not; 2) Position, to detect the location of cracks; 3) Segmentation, to identify distribution, topology and size of cracks, which can be divided to patch-level and pixel-level according to segmentation fineness.

‾ Shengxia Road NO.666, Pudong, Shanghai 201210, P.R. China

Hand-crafted feature engineering includes: 1) Edge/morphology detection algorithms: e.g., Canny, Sobel, histogram of oriented gradient (HOG), local binary pattern (LBP) and 2) Feature transformation algorithms: e.g., fast Haar transform (FHT), fast Fourier transform (FFT), Gabor filters, intensity thresholding. These algorithms don't have to learn from data set, and most of the mathematical calculations are analytical and fast with lightweight computation. The shortcomings are the weak generalization ability to various random variable factors. Once the application scenario or environment changes, parameters need to be fine-tuned, or the algorithm needs to be redesigned and even fails at all.

Both machine learning and deep learning are learning methods depending on data set. The difference is that the mathematical expression of machine learning is explicit and explicable, while deep learning is implicit. Common crack identification machine learning algorithms include CrackIT[7], CrackTree[8], CrackForest[9], etc. It is generally believed that deep learning is oriented towards higher-dimensional features, while machine learning is towards lower-dimensional features. Machine learning is somewhere between hand-crafted feature engineering and deep learning methods which automatically extract high-dimensional features from large-scale data sets, with strong generalization ability and high accuracy. Deep learning has continuously obtained State of the Art (SOTA) progress in various missions. However, it still needs great efforts on theoretical research and visualization technology to solve the problems of interpretability known as "Black Box" issue[28-30].

This paper focused on new progress of three categories of deep learning methods, i.e., *Transfer Learning* (TL), *Encoder-Decoder* (ED) and *Generative Adversarial Networks* (GAN), in pavement (concrete) crack image identification. The performance and effect of different neural network models of certain algorithm as well as algorithms of various methods were compared and evaluated within and between categories. The contributions of this work include:

- The fundamental frameworks and basic characteristics of TL, ED and GAN are presented. The recent developments and progresses of these algorithms on pavement crack identification as well as the evolution of CNN backbone models and GAN models are summarized. Common architecture, common modules, and specific techniques for improving performance of typical crack identification models are highlighted.
- The patch sample classification performance, full-size image segmentation performance and detection effect were tested on pavement crack public data sets such as SDNET2018, CFD. The performance discrimination of different evaluation metrics on crack identification was compared. Then a quantitative evaluation indicator is also advised for the situation of crack GT semi-accurate (e.g., 1-pixel curve) labeling.
- A weakly supervised learning framework combining TL and semi-supervised GAN is proposed, which can maintain comparable crack identification performance with the supervised learning while greatly reducing the number of labeled samples used, through the measures of utilizing fine-tuned TL backbone model, controlling the ratio of labeled samples and unlabeled ones, and adding additional unlabeled samples.

## 2. Related Works

### 2.1. *Transfer Learning (TL)*

TL uses CNN backbone networks to learn reusable basic features. These backbone networks are usually SOTA models that had been architecture-optimized and pre-trained on large-scale generic data sets. Then the weight parameters of the upper and/or output layers are fine-tuned on specific data set (e.g., pavement crack data set), to learn higher-dimensional features. TL is suitable for small data set, easy to adjust, with good generalization performance and the training is fast.

#### 2.1.1. *CNN backbone networks*

The *evolution* of common CNN backbone networks is demonstrated in Fig.1. Tab.1 lists the memory size, and Top-1/Top-5 accuracy on the ImageNet validation set of typical backbone network models when transfer learning.

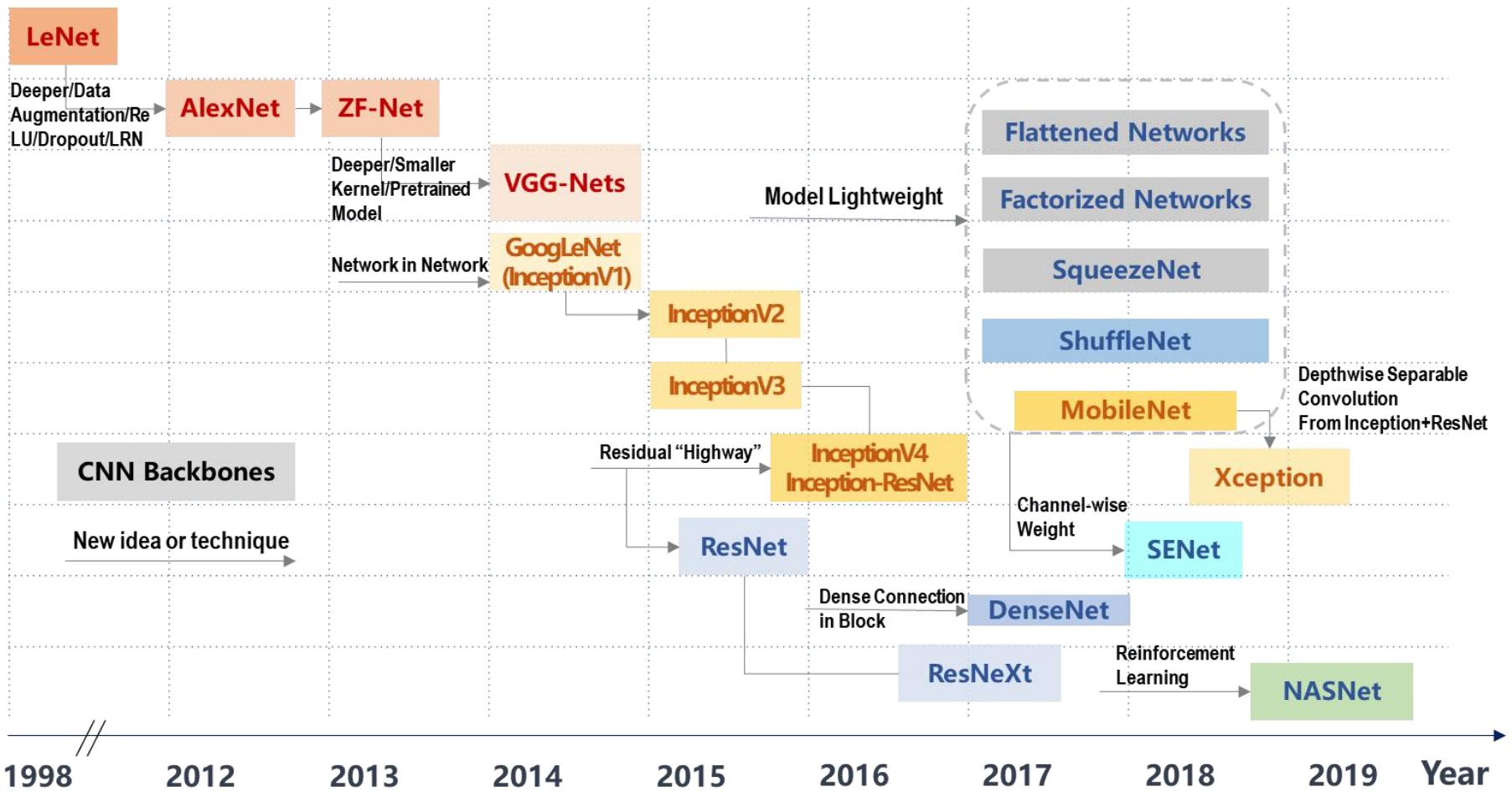

Fig. 1. Evolution of CNN backbone networks since LeNet.

Table 1. Typical CNN backbone models for transfer learning.

| Model | Size (MB) | Top-1 Acc. | Top-5 Acc. | Model | Size (MB) | Top-1 Acc. | Top-5 Acc. |
|---|---|---|---|---|---|---|---|
| VGG16 | 528 | 0.713 | 0.901 | InceptionResNetV2 | 215 | 0.803 | 0.953 |
| VGG19 | 549 | 0.713 | 0.900 | MobileNetV1 | 16 | 0.704 | 0.895 |
| ResNet50V2 | 98 | 0.760 | 0.930 | MobileNetV2 | 14 | 0.713 | 0.901 |
| ResNet101V2 | 171 | 0.772 | 0.938 | DenseNet121 | 33 | 0.750 | 0.923 |
| ResNet152V2 | 232 | 0.780 | 0.942 | DenseNet201 | 80 | 0.773 | 0.936 |
| InceptionV3 | 92 | 0.779 | 0.937 | Xception | 88 | 0.790 | 0.945 |

### 2.1.2. *Framework and procedure of TL*

The overall framework and procedure of deep TL[3], including data flow and step flow, is shown in Fig.2. The main steps are as follows: 1) Add your custom layers on top of a pre-trained base network. The base network is composed of a backbone network model and its weights pre-trained over a large data set. 2) Freeze the base network. 3) Train the added top layers on specific data sets. 4) Unfreeze some layers in base. 5) Jointly train both unfreezed layers and the top layers added. Steps 2) to 5) are collectively known as fine-tuning. Finally, if the preset target is met, the new network model and weight parameters fine-tuned will be saved for deploying and used to test samples to be predicted. After predicting, the probability value and class label will be the output. If not, the above steps can be redone to optimize the network model and weight parameters until the target is met, through increasing sample number, re-selecting backbone network and unfreezing more layers etc.

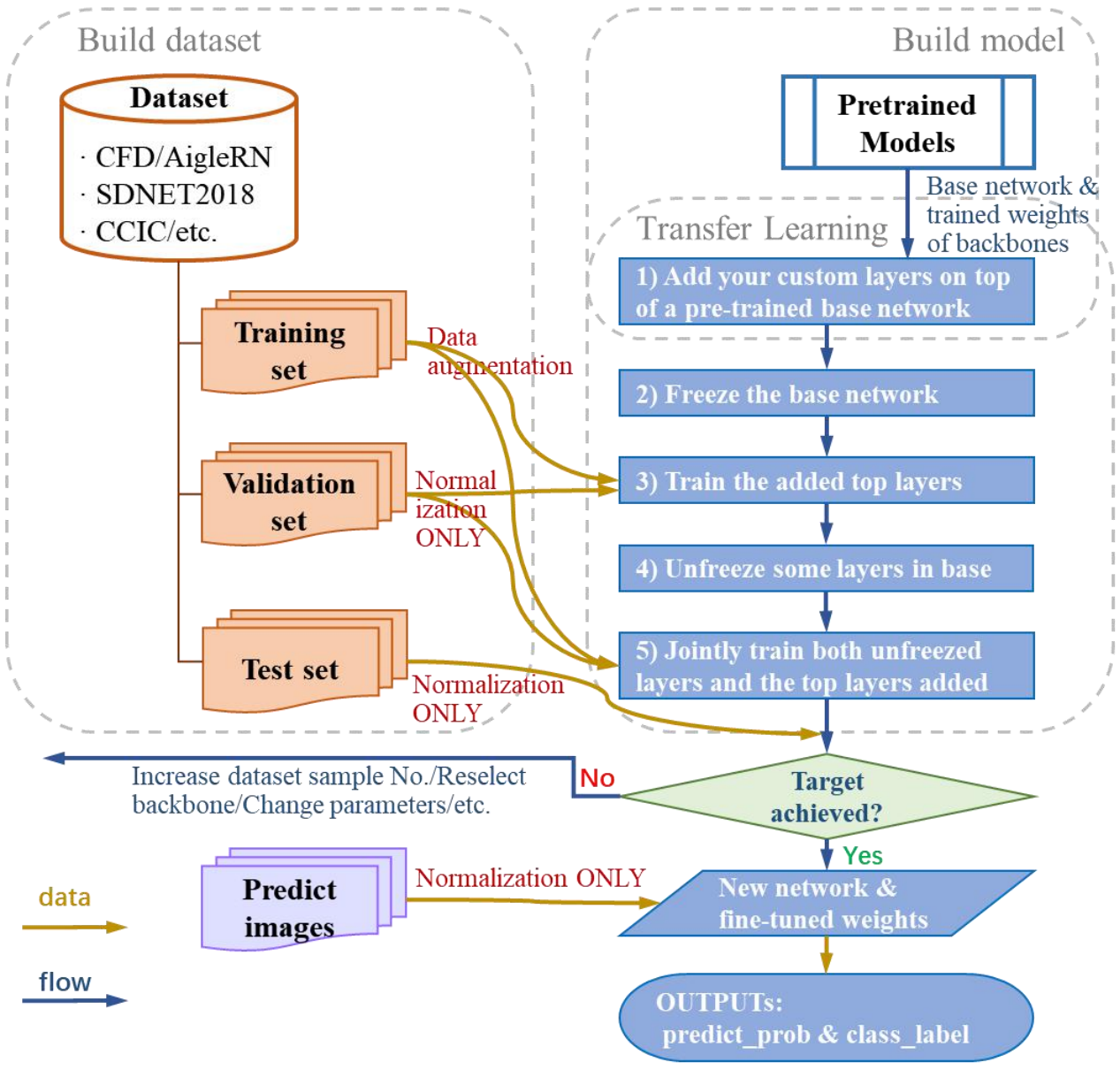

Fig. 2. Framework and process of deep TL.

## 2.2. *Encoder-Decoder (ED)*

### 2.2.1. *Motivation*

ED, as the name implies, consists of an encoder and a decoder. In CNN, the role of encoder network is to produce feature maps with semantic information. The role of decoder network is to map the low-resolution features output by the encoder back to original size of the input image for per-pixel classification. ED is suitable for unsupervised/weakly-supervised and small data set situations.

ED framework can make up the shortcomings of CNN[25]/FCN[19]algorithms on identifying complex crack topology (e.g., alligator crack), i.e., 1) Pavement cracks have various morphology and topology. However, the receptive field of the CNN/FCN filters to extract features is a kernel of specific size (especially 3×3, 5×5, 7×7), which limits the range and robustness to crack detection. 2) It has not been considered that crack edges, patterns or texture features contribute differently to detection results.

### 2.2.2. *Architecture of typical ED models*

ED based FPCNet[1] is one of SOTA models with excellent accuracy and speed, including two sub-modules: *multi-dilation* (MD) module and SE-Upsampling (SEU) module. Another model U-HDN[10], similar to FPCNet[1], integrated multi-dilation (MD) module and *hierarchical feature* (HF) learning module on the basis of U-net[11] architecture. The compared two models are composed of similar or common sub-modules, such as MD module (Fig.3) based on dilated convolution kernel operation, SEU module (Fig.4), U-net-likewise ED main architecture module (Fig.5, Fig.6). The dilated convolution enlarges the kernel's *context* window size, instead of using subsampling operation or a larger filter with much more parameters.

As shown in Fig.3, the MD module concatenates 6 branches, i.e., four dilated convolutions (double operations per branch) with rates of {1, 2, 3, 4} (or {1, 2, 4, 8} / {2, 4, 8, 16}, can be set according to the statistics of crack width), a global pooling layer and the original crack multiple-convolutional (MC) features. The latter two branches in the dashed rectangular box in Fig.3, can also be integrated into the ED main architecture module as designed by U-HDN[10]. After the concatenation, a 1×1 convolution is performed to obtain the crack MD features. By concatenating these feature extraction branches, the MD module can extract context features ranging from pixel level to global level, thus having the ability to detect cracks with different widths and topology. In the figure, H, W and C denote the height, width and channels of feature maps respectively.

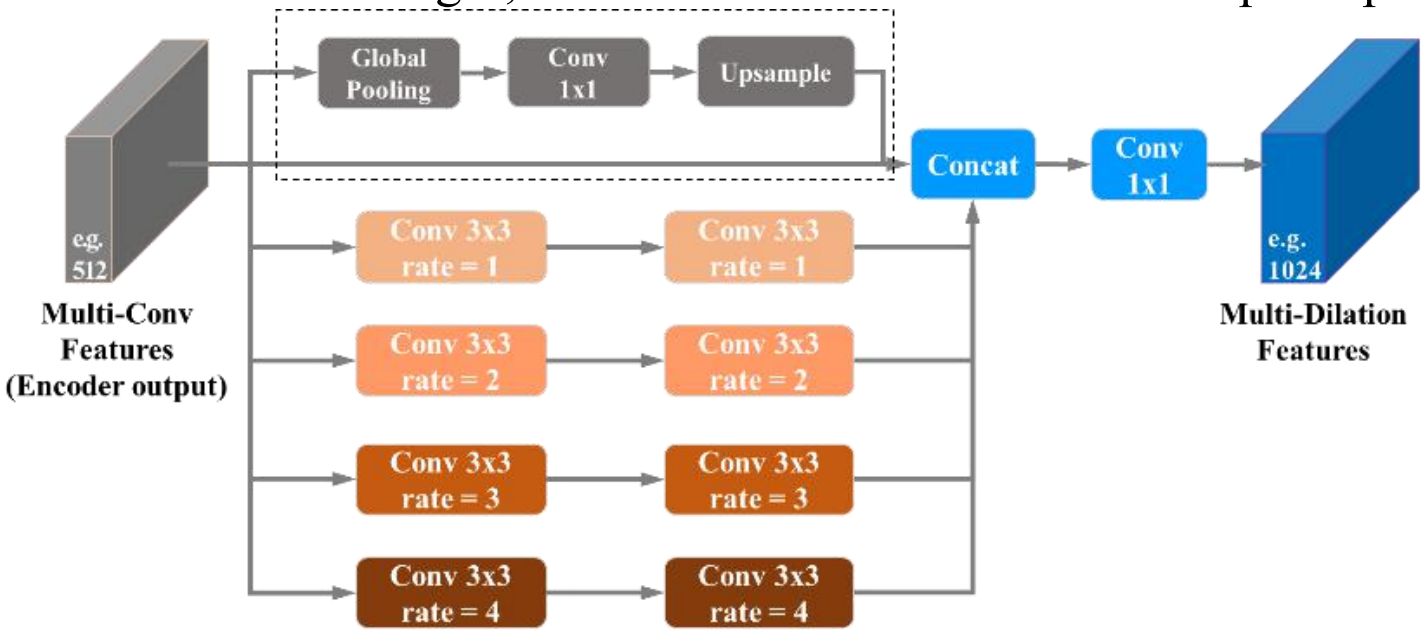

Fig. 3. Multi-Dilation module (reproduced from Ref.1,10).

The purpose of the upsampling module is to restore the resolution of MD features to that of the original input image to realize pixel-level crack identification. The SEU module of FPCNet[1], as shown in Fig.4, performs up-sampling operations (e.g., transposed convolution or bilinear interpolation) in decoder to continuously restore the resolution of the MD feature, meanwhile, the squeezing and excitation (SE) module is integrated. The inputs of SEU module are MD features and MC features, and the output is the optimized MD features after weighted fusion. The implementation details are as below.

(1) The SEU module first restored the resolution of the crack MD features through transposed convolution. Then, it added the MC features to the MD features in order to fuse the associated crack information concerning the edge, pattern, texture among others.

(2) Subsequently, the SE operation was applied to the added MD features to learn the weights of different features. Global pooling was first performed to obtain the global information of the C channels. After squeeze ($F_{sq}$) and excitation ($F_{ex}$) (two fully-connected layers) of the global information, the weight of each feature for its channel was obtained. Through the SE learning, the SEU module could adaptively assign different weights to different crack features such as the edge, pattern, and texture.

(3) Finally, each feature in the added MD features was multiplied ($F_{scale}$) by its corresponding weight to obtain the optimized MD features.

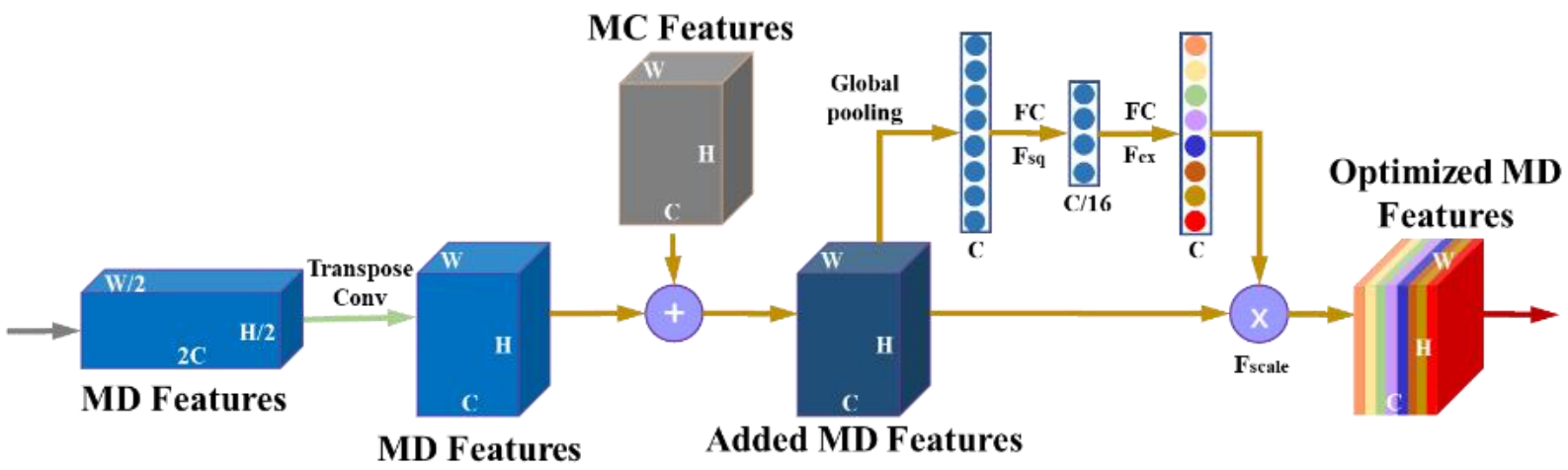

Fig. 4. SE-Upsampling module (reproduced from Ref.1).

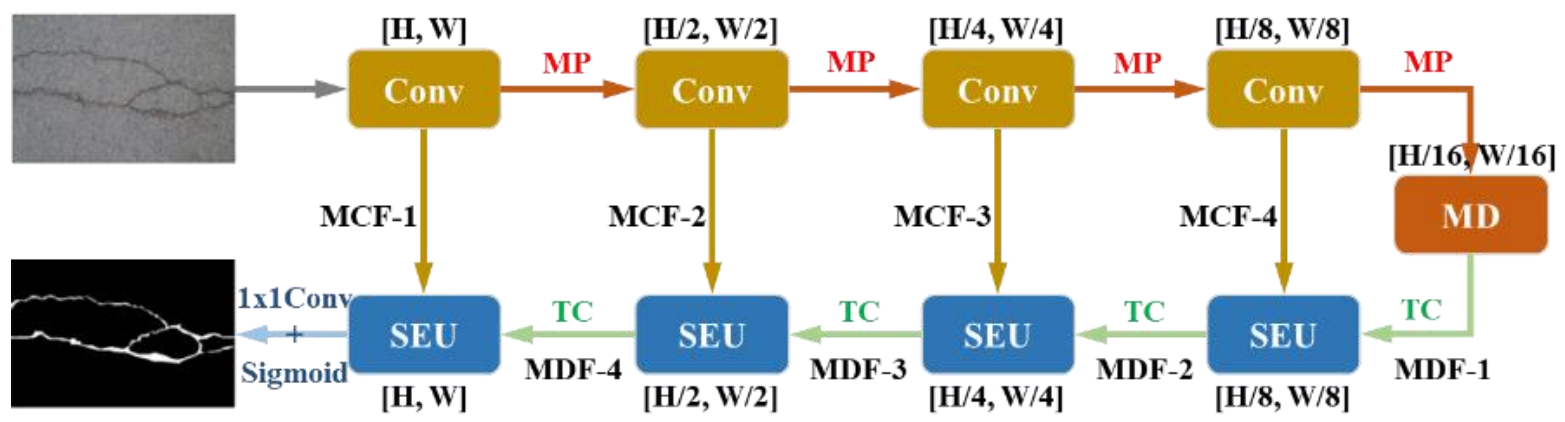

Fig. 5. Network overall architecture of FPCNet (reproduced from Ref.1).

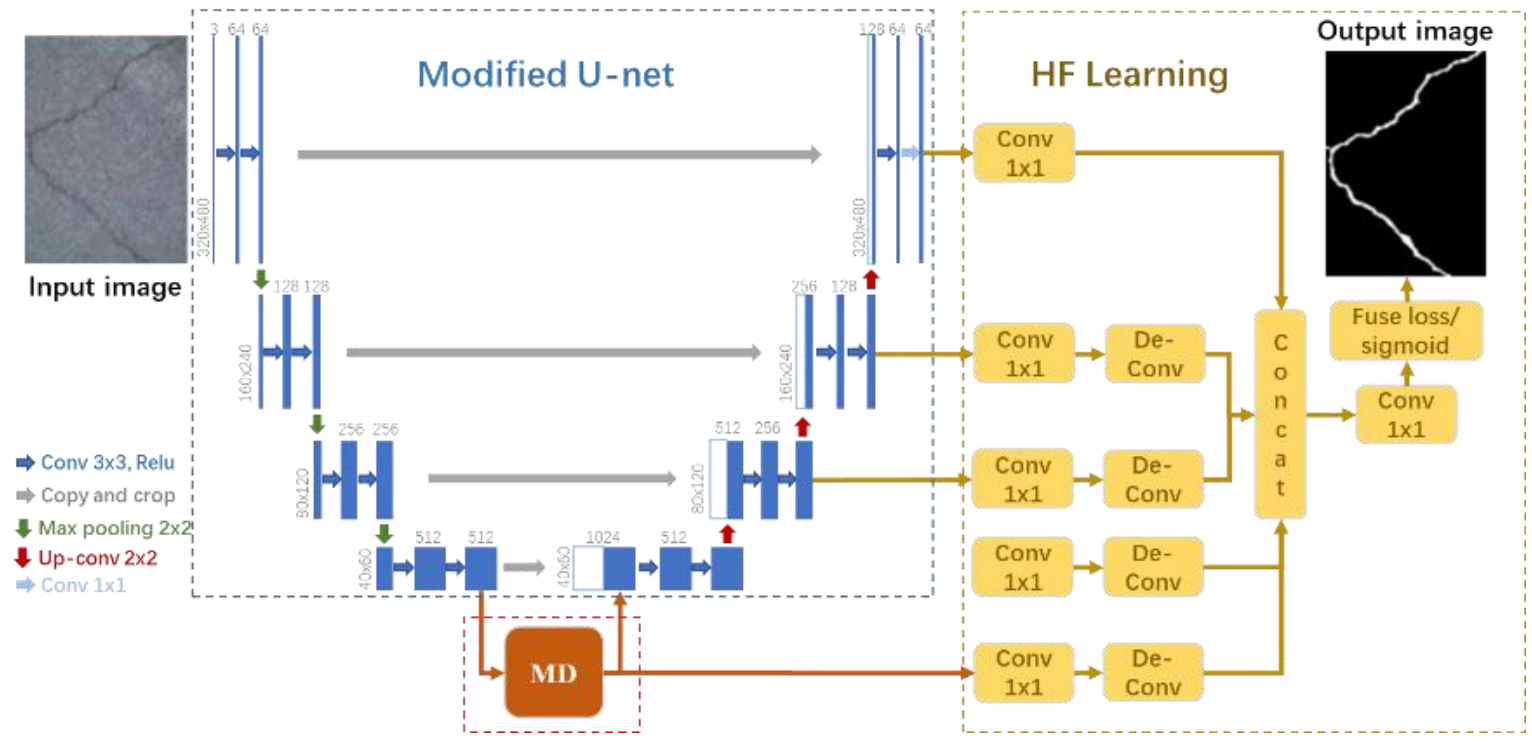

Fig. 6. Network overall architecture of U-HDN (reproduced from Ref.10).

The overall model architectures of FPCNet[1] and U-HDN[10] are illustrated in Fig.5 and Fig.6 respectively. FPCNet[1] embedded the above MD module and SEU module into common framework ED in semantic segmentation network. It used 4 Convs (two 3×3 convolutions and ReLUs) + max poolings as the encoder to extract features. Next, the MD module was employed to obtain the information of multiple context sizes. Subsequently, 4 SEU modules were operated as the decoder. In Fig.5, H and W indicate the original sizes of the image. The red, green, and blue arrows indicate the max pooling, transposed convolution and 1×1 convolution + sigmoid, respectively. MCF denotes the MC features extracted in the encoder, and MDF denotes the MD features. U-HDN[10] integrated the MD module (bounded in red dotted rectangular box in Fig.6) and the HF learning module (in yellow dotted box in Fig.6), based on the modified U-net architecture (in blue dotted box in Fig.6), namely, layers such as Pool4, Conv9, Conv10 and Up-conv1 were removed, and zero filling was adopted during the up-convolution and down-convolution paths.

In addition, Li et al.[13], inspired by DenseNet, fused the densely connected convolution module and the deep supervision module to extract more detailed features of cracks. Aiming at the data imbalance between cracks and background samples, a class-balanced cross-entropy loss function was designed to obtain more stable training results by dynamically adjusting the weight of crack pixel loss. Yu et al.[14] introduced Capsule networks to develop CCapFPN, which is a novel capsule feature pyramid network architecture. It included a bottom-up pathway functioned to extract different levels and different scales of capsule features, a top-down pathway together with lateral connections to fuse the capsule features to provide high-resolution, semantically strong feature representations, and a context-augmented module was added to enlarge the receptive field. By these modules, CCapFPN[14] achieved a good crack segmentation performance. In summary, the above-mentioned ED models contribute common idea, similar architecture and same fundamental modules.

## 2.3. *Generative Adversarial Networks (GAN)*

### 2.3.1. *Overview*

GAN make the generated samples obey the probability distribution of real data via two networks' (namely, discriminator and generator) adversarial training. GAN has many advantages such as accurate estimation of density function, efficient generation of required samples, elimination of deterministic bias and good compatibility with various neural network structures. Therefore, it has received extensive attention from the research community. It has been combined with various architectures, algorithms and techniques of neural networks to solve the problem that GANs are difficult to train and evaluate (especially concerned the quality and diversity of generated images). Thus, a variety of GAN variants were derived, which enables and extends the application of neural network in Computer Vision[15-17].

According to Ref.15, GAN variants can be divided into two categories: architecture-variant and loss-variant. The former category includes subcategories such as network architecture, latent space, application focused, etc. The latter includes loss type (IPM/non-IPM based), regularization etc. The evolution diagram and main characteristics of common GAN models are shown in Fig.7.

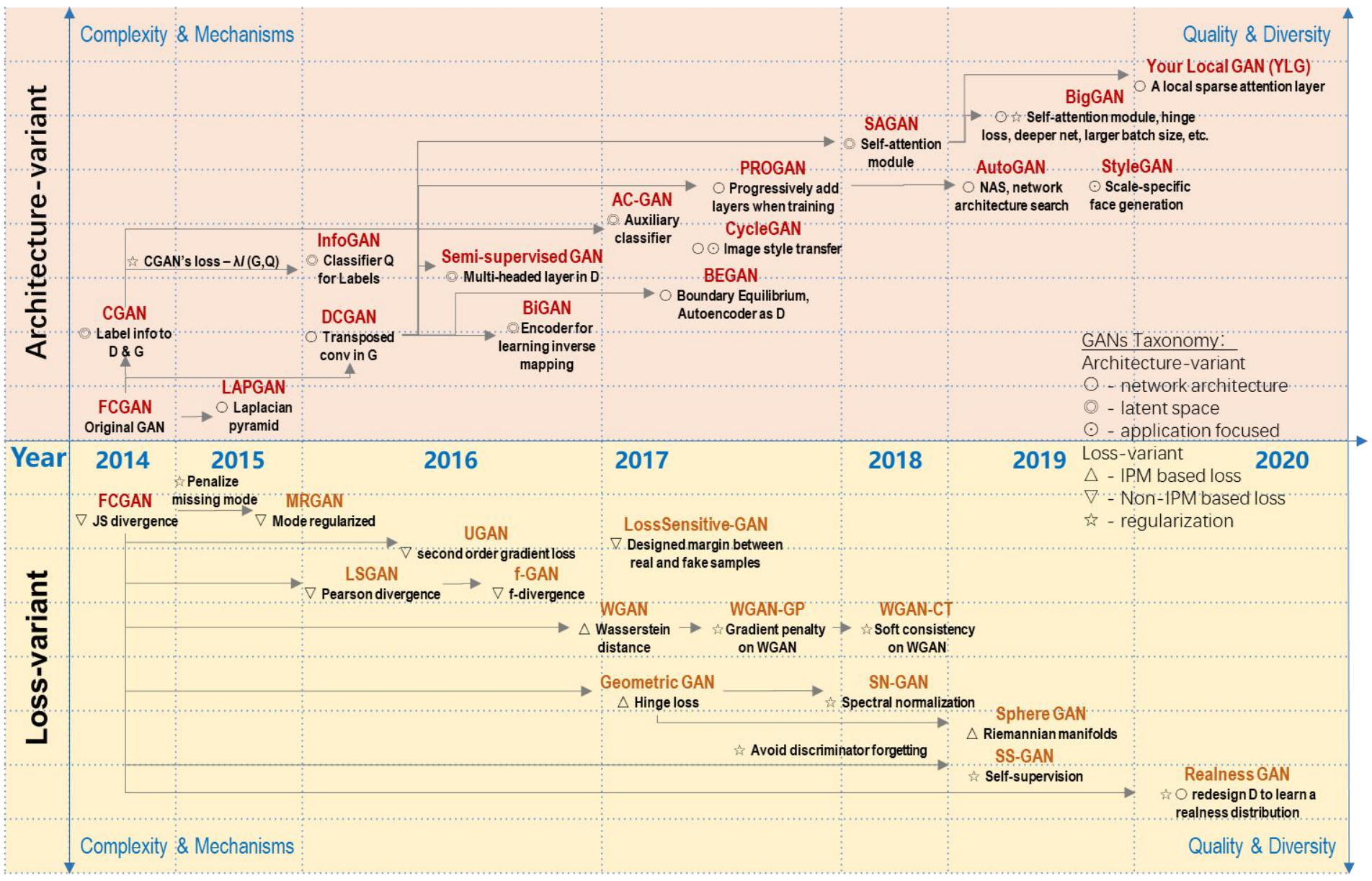

Fig. 7. Evolution of GAN Variants (reproduced from Ref.15-17).

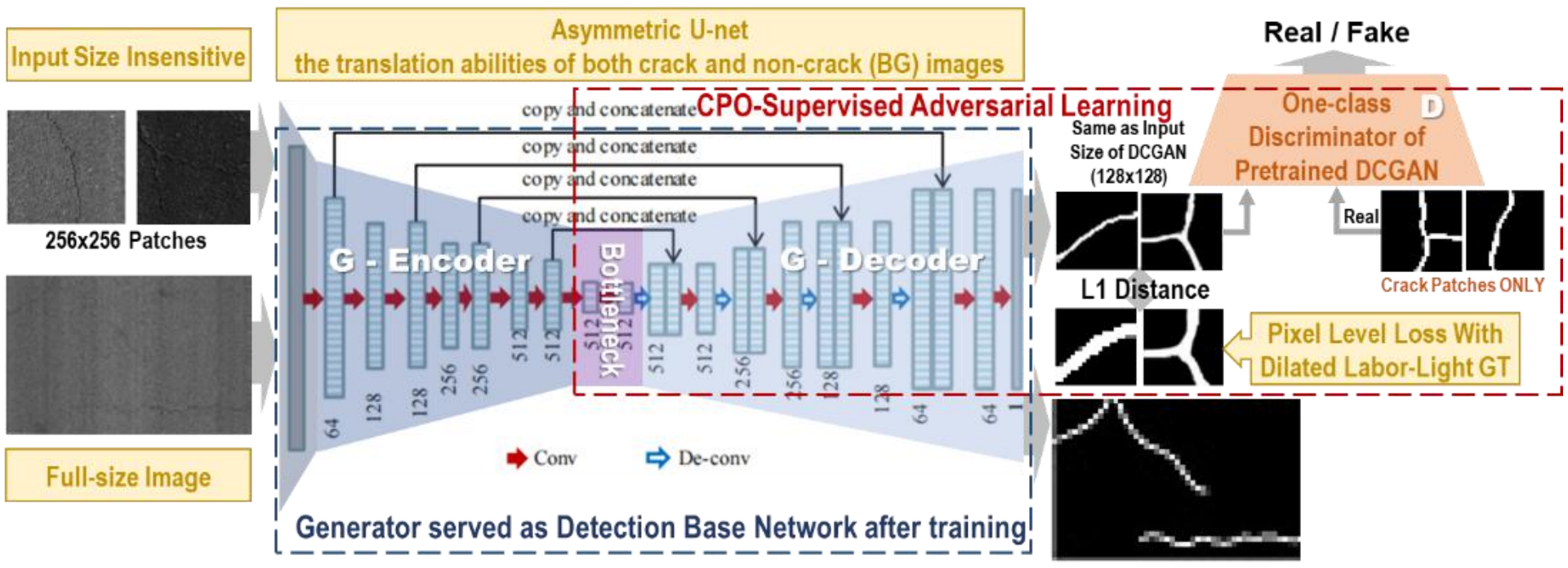

Remarks: a) D is a One-class DCGAN discriminator that is pre-trained only by crack-GT patches, which will enable the network to always generate crack-GT images. It is a key module to overcome the "All-Black" problem. b) Pixel-level loss is used to ensure that the generated crack pattern is the same as the input patches, which is achieved by optimizing the L1 distance from Dilated Labor-light GT, for example 1 pixel dilated 3 times. The total loss is L1 distance added to adversarial loss generated by the pre-trained DCGAN. c) Because the network is only trained with crack patches, an asymmetric U-net network is introduced to realize the translational interpreting ability of crack and non-crack images (including background images) with severely unbalanced sample numbers. d) After the training is completed, the generator G will act as the crack detection network. e) The whole network is a fully convolutional network (FCN), which can process arbitrary full-size crack images.

Fig. 8. Architecture overview of CrackGAN (reproduced from Ref.2).

### 2.3.2. *Progress in GAN algorithms for crack identification*

GAN variants are well combined with the architectures, algorithms and techniques of various neural networks, and can also be applied for hard samples mining as well as semi-supervised/unsupervised learning.

ConnCrack[18] and CrackGAN[2] are representative GAN models for crack identification, with high accuracy. However, the training and prediction time was long in earlier version. The improved CrackGAN[2], based on previous GAN models such as DC-GAN and with encoder-decoder as the generator, proposed the *crack-patch-only* (CPO) supervised adversarial learning and the asymmetric U-Net architecture to perform end-to-end training with partially accurate GTs (i.e., 1-pixel curve manual labeling) generated by *labor-light* manner. It can reduce the workload of GT labeling significantly and achieve great performance when dealing with full-size images on pixel-level crack segmentation. The proposed approach also employed transfer learning to train the prototype of the encoding network, and transferred the knowledge from a pre-trained DC-GAN to provide the generative adversarial loss for the end-to-end training. The overall architecture of CrackGAN[2] is shown in Fig.8.

CrackGAN[2] for the first time solved a practical and essential problem named "All Black" issue, i.e., the network converged to the state that the whole crack image was regarded as the background (labeled as 0). It is a common problem that both CNN/FCN and ED models may encounter, which is caused by 1) The data imbalance of crack samples and background uncrack samples, and 2) Blurred boundaries of tiny long cracks that per-pixel accurate labeling is difficult or infeasible. And the computational efficiency is greatly improved (predicting a 2048×4096-pixel full-size image takes about 1.6s on NVIDIA 1080Ti GPU); while the labeling workload significantly reduced. To be best of our knowledge, it is currently the SOTA architecture or model. The design ideas can be referred as design guideline to solve the problem of fuzzy boundary in image segmentation, and is expected to realize real-time detection of pavement cracks. For more implementation details, please refer to the original paper Ref.2.

## 3. Evaluation and Comparison of the Crack Identification Algorithms

### 3.1. *Public Crack Datasets*

The typical pavement (concrete) crack public data sets and their samples' features, download sources were collected and listed in Ref.3. The sample images in the data sets were all captured by personal cameras, mobile phone cameras, car or industry cameras, thus acquisition cost is low. The number of samples varies from dozens to thousands depending on the purpose and requirement of the data set, e.g., classification or segmentation, training or testing. And the resolution varies with acquisition device. According to the needs of subsequent algorithms, original images can be cropped into small patches, such as 256x256 or 224x224. The sample number can also be further increased via data enhancement. Although the pavement (concrete) crack is taken as the use case, the deep learning algorithms discussed in this paper can be easily transplanted to crack identification or segmentation in other engineering structures and even other image classification/segmentation application scenarios.

### 3.1.1. *SDNET2018*

SDNET2018 is a patch-level annotated data set for training, validation and benchmark test of crack detection artificial intelligence algorithms. The original images were taken with a 16 MP Nikon digital camera. It includes 230 photos of cracked and uncracked concrete surfaces (54 bridge decks, 72 walls, and 104 sidewalks). Each photo is then cropped into 256×256 patches. If there is a crack in the sample, it is marked as C; if not, marked as U. In total, SDNET2018 contains 56,092 sample images. The crack size in the positive samples is as narrow to 0.06 mm and as wide to 25 mm. Random variable factors such as environmental influence and background interference obstacles are also included, as shown in Table 2, many positive and negative samples in the data set are difficult to be recognized by human eyes.

Table 2. Random variable factors in SDNET2018.

<table>
<tr><th colspan="2" rowspan="2">Random Variable Factors</th><th colspan="2">Sample Patches</th></tr>
<tr><th>Negative Samples</th><th>Positive Samples (With Crack)</th></tr>
<tr><td colspan="2">Environmental:<br>Shadow, illumination change, low contrast</td><td>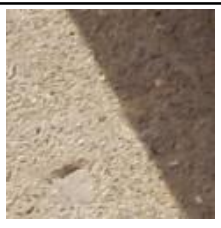</td><td>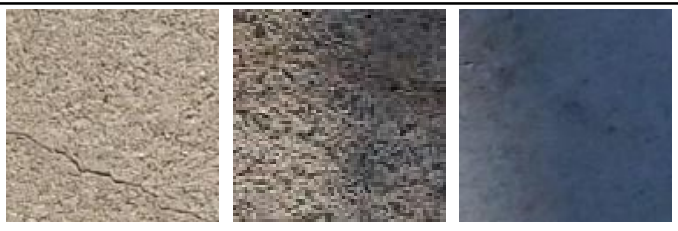</td></tr>
<tr><td colspan="2">Background:<br>Oil/Wet stain, Texture, Spots, Surface roughness</td><td>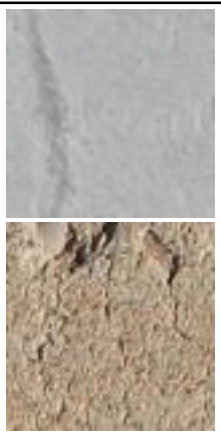</td><td>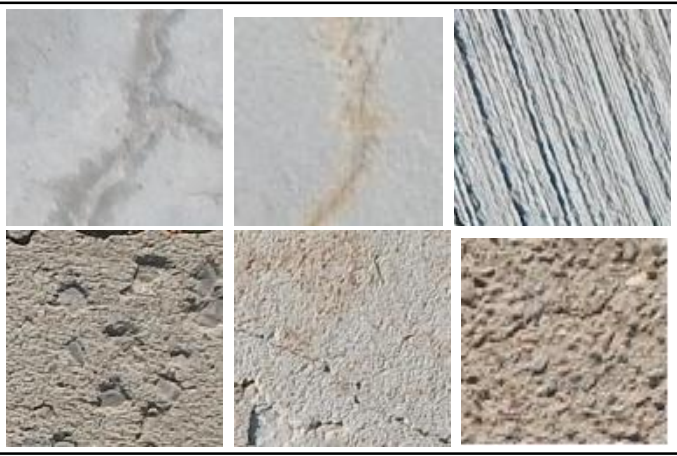</td></tr>
<tr><td rowspan="2">Interference</td><td>Holes, Process gaps, edge</td><td>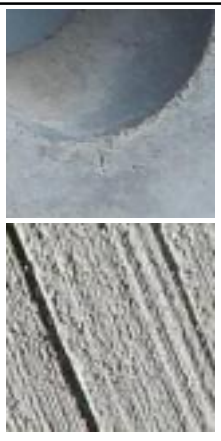</td><td>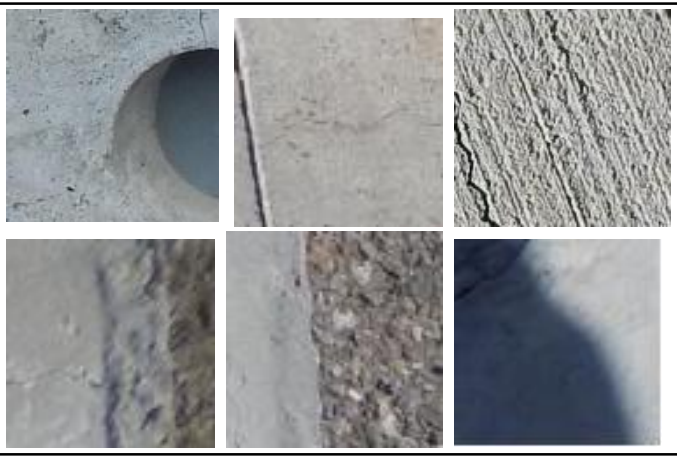</td></tr>
<tr><td>Cracking wood, Threads, Weeds</td><td colspan="2">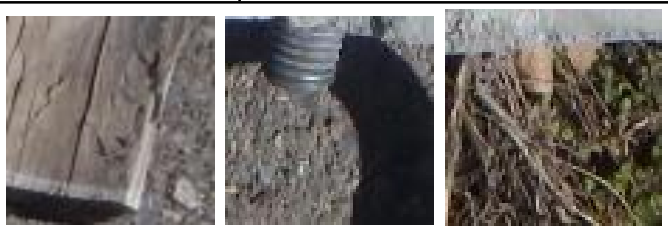<br>(This line are all negative samples without crack)</td></tr>
</table>

### 3.1.2. *Crack forest dataset (CFD)*

CFD is a pixel-level annotated pavement crack data set reflecting the overall situation of Beijing urban pavement. It is one of the benchmark baseline data sets. A total of 118 images were collected using a mobile phone (iPhone 5). The images contain noise or interference factors such as lane lines, shadows, oil stains, etc.

## 3.2. *Crack Classification Evaluation*

### 3.2.1. *Classification benchmark test results and discussion*

A transfer learning performance benchmark test was implemented on SDNET2018, selecting MobileNetV2, InceptionV3, Resnet152V2 and InceptionResnetV2 as backbone network respectively. Class weight was used to balance data imbalance. The patch classification results of TL algorithms are compared with FCN[19] (FCN is a one-stage pixel-level semantic segmentation method that does not require window sliding) and the ED model FPCNet[17], as shown in Table 3.

Table 3. Classification test results and performance comparison of typical backbone models on patches.

| Algorithm and model [a)] | Accuracy | Precision | Recall | $F_1$-score | Predicting Time @ GTX1080Ti [b)] |
|---|---|---|---|---|---|
| TL-MobileNetV2 | 0.936 8 | 0.947 7 | 0.973 2 | 0.960 3 | 8.1 ms/patch |
| TL-InceptionV3 | 0.941 0 | 0.952 4 | 0.980 6 | 0.966 3 | 16.1 ms/patch |
| TL-Resnet152V2 | 0.956 1 | 0.961 3 | 0.981 7 | 0.971 4 | 50.2 ms/patch |
| Original FCN[19] | 0.965 8 | 0.972 9 | 0.945 6 | 0.959 0 | 19.8 ms/patch |
| ED-FPCNet[17] | 0.970 7 | 0.974 8 | 0.963 9 | 0.969 3 | 67.9 ms/patch |

Notes: a) TL algorithms were tested on SDNET2018, while FCN and ED algorithms were on CFD. The results are average of 10 tests. CFD is relatively more recognizable than SDNet2018. b) Predicting time includes image loading and pre-processing time as well as inference time. The resolutions of input patch samples to TL models were 224x224 and 229x229(InceptionResnetV2), and those of FCN/ED models were 256x256/288x288.

It can be seen from Table 3 that:

(1) Using TL method to classify crack or uncrack patch images, the testing accuracy on SDNET2018 which is difficult for human eyes to identify, has exceeded the ImageNet multi-classification baseline (seen in Table 1).

(2) The fine-tuned TL algorithm is close to or even slightly better than (e.g., Recall and $F_1$-score index) the classification test accuracy of FCN and ED algorithms, which is attributed to the fact that the backbone network pre-trained by large-scale dataset ImageNet and optimized in architecture can better extract high-dimensional general features. Thus, the predicting time of TL's 8.1~50.2ms is much lower than that of FPCNet[17], which is 67.9ms (all @ NVIDIA GTX1080Ti GPU platform).

The TL algorithms are usually trained and tested on the patch samples cropped from the original images. Due to the limitation of the minimum size of the patches, it is generally used for classification problem and has poor segmentation effect on full-size crack image (as shown in Fig.10 the 3rd line produced with RCNN-based Algorithm[20]). The segmentation problem mainly depends on ED and GAN algorithms. On the other hand, although the imbalance of crack and uncrack/background sample size (SDNET2018 is about 1:5.6) can be compensated by the class weight or class balanced loss function. However, due to the ambiguity of crack boundary, it is difficult to accurately per-pixel annotate by human (semi-accurate 1-pixel curve is usually manually annotated, that is, 2 pixels labeling error). Pixel level mismatching caused by inaccurate GT annotation may cause to encounter the "All-Black" problem. Moreover, it is worth noting that this also leads to the possibility that even models with higher scores on accuracy, precision, recall, or $F_1$-score, may not necessarily perform better in actual pixel-level segmentation. Therefore, new evaluation index needs to be introduced.

### 3.3. *Evaluation Index HD-score for Crack Segmentation*

The Hausdorff Distance score (HD-score) is a mathematical construct to measure the "closeness" of two sets of points that are subsets of a metric space. Such a measure can be used to assign a scalar score to the similarity between two trajectories, data clouds or any sets of points.

As illustrated in Fig.9, for two point sets $P$ and $Q$, Hausdorff Distance is defined as

$$H(P,Q) = \max\left[h(P,Q),\ h(Q,P)\right]. \tag{1}$$

Where,

$$h(P,Q) = \max_{p\in P} \min_{q\in Q} \|p - q\|. \tag{2}$$

The penalty function is defined as

$$h_{\text{penalty}}(P,Q) = \frac{1}{|P|}\sum_{p\in P} \text{sat}_u \min_{q\in Q} \|p - q\|. \tag{3}$$

Where, parameter $u$ is the upper limit of the saturated function **sat**, which is used to directly remove false positives (FP) far from GT.

Suppose $P$ is the identified crack boundary pixel point set and $Q$ is the point set of ground truth (GT), then the HD-score of the crack boundary point sets is

$$HD\text{-score}(P,Q) = 100 - \frac{H_{\text{crack}}}{u} \times 100. \tag{4}$$

Where,

$$H_{\text{crack}}(P,Q) = \max\left[h_{\text{penalty}}(P,Q),\ h_{\text{penalty}}(Q,P)\right]. \tag{5}$$

The HD-score is sensitive to the image segmentation boundary and insensitive to the inherent foreground (crack) and background imbalance in the long-narrow object (crack) detection, so it is suitable for evaluating the positioning accuracy of crack boundary segmentation[2, 22].

### 3.4. *Performance and Effect of Crack Segmentation*

The test results on CFD are shown in Table 4 and Fig.10. Ref.20 proposed a patch-based RCNN method, which cannot accurately locate the crack location due to the limitation of patch size. And the prediction of full-size images requires using sliding window, making prediction time much longer. It takes about 10.2s for a single CFD sample. FCN-VGG[23] is a pixel-level identification algorithm, via end-to-end training of accurate GT annotation at each pixel. When there is a deviation in GT, it fails to detect thin cracks. DeepCrack[24] achieves good results with multi-scale hierarchical fusion, but it also relies on accurate GT. As mentioned above (shown in Fig.8), by introducing CPO supervision and asymmetric U-net, CrackGAN[2] proposed the GAN architecture with a single-class discriminator. It treated all-black background patch as a false sample, avoiding the "All-Black" problem caused by the inherent imbalance of positive and negative sample data in crack identification, and significantly improved the ability of accurate crack segmentation, with HD-score as high as 96, especially for the identification of thin cracks. Furthermore, it also has high computational efficiency, predicting a full-size CFD image only takes about 1.6s at NVIDIA 1080TI GPU.

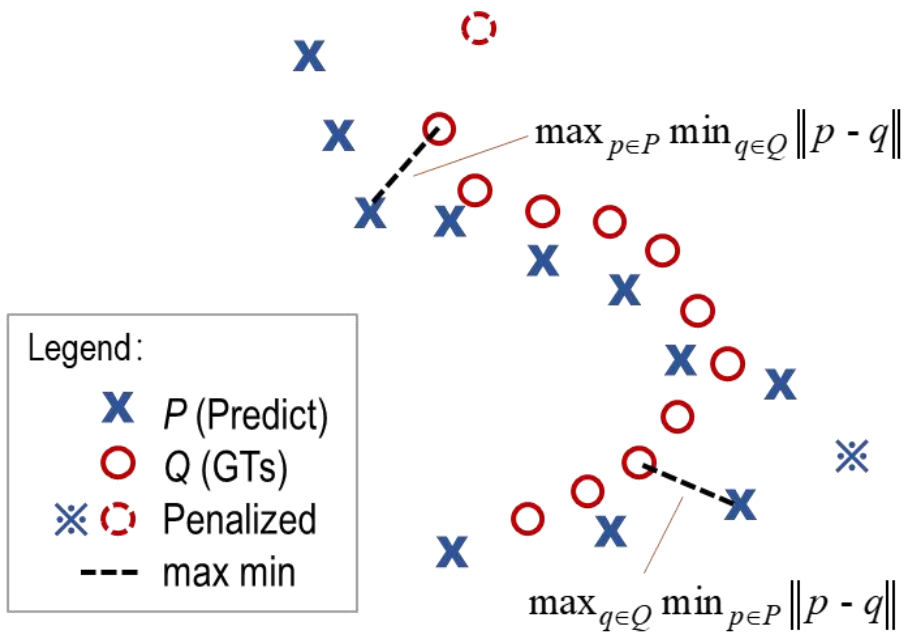

Fig. 9. Illustration of the Hausdorff Distance.

Table 4. Segmentation performance comparison on CFD.

| Method | HD-score | Prediction Time (s) |
|---|---|---|
| RCNN-based[20] | 70 | 10.2 |
| FCN-VGG[23] | 88 | 2.8 |
| DeepCrack[24] | 94 | 2.4 |
| CrackGAN[2] | 96 | 1.6 |

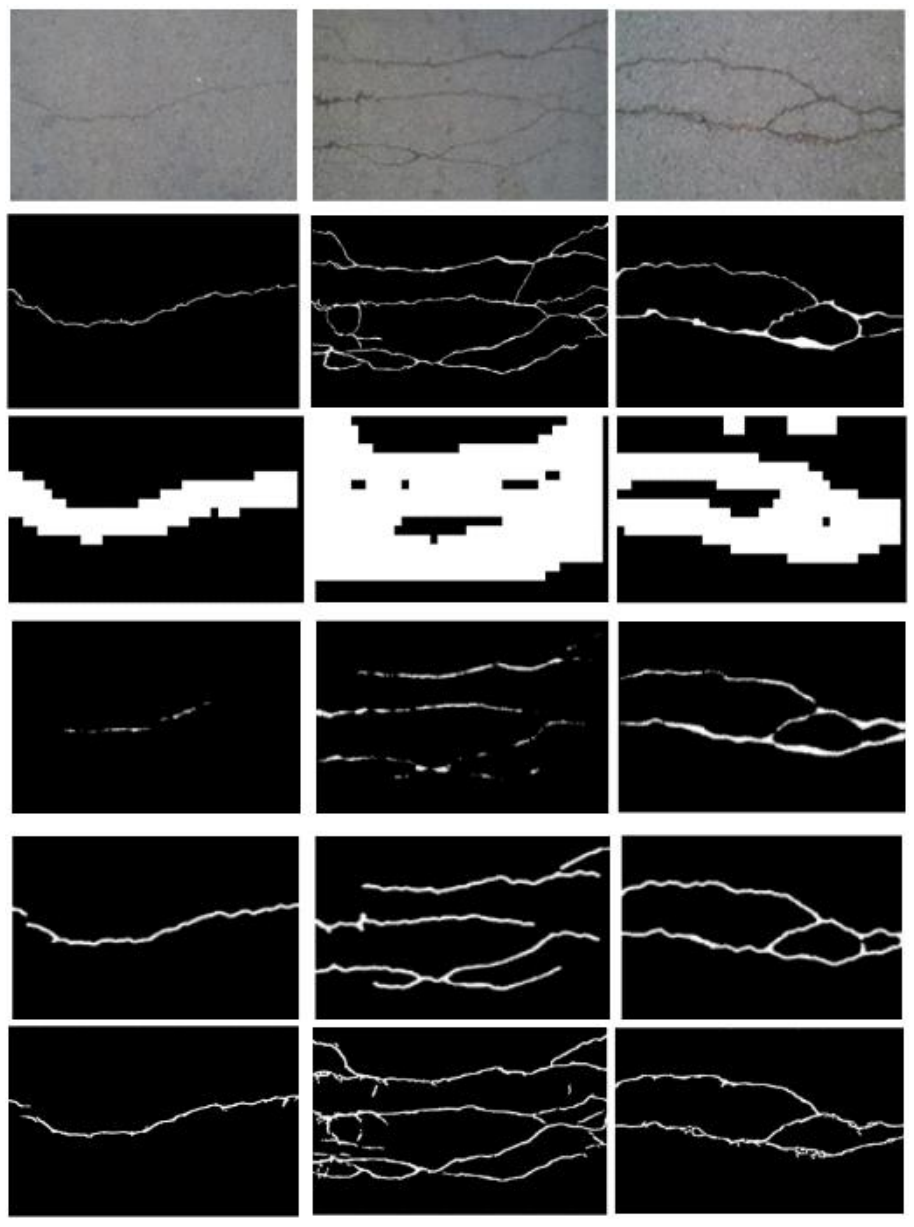

(From top to bottom: Original image, Ground Truth, RCNN-based[20], FCN-VGG[23], DeepCrack[24], CrackGAN[2])

Fig. 10. Comparison of detection results on CFD[2].

To the best of our knowledge, CrackGAN[2] is the currently published SOTA model. Meanwhile, the HD-score quantization value has a strong correlation and matches well with the crack segmentation effect, thus can be used as the quantitative evaluation index of crack identification performance, where GT per pixel accurate labeling is difficult or not feasible.

### 3.5. *TL-SSGAN Weakly Supervised Learning Framework and Evaluation Test*

CrackGAN[2] utilized the semi-accurate labor-light annotation method of 1-pixel curve manual labeling to greatly reduce the labeling difficulty and workload. On the other hand, semi-supervised GAN (SSGAN) can also greatly reduce the need for labeled samples from the perspective of semi-supervision[26,27]. A *weakly supervised* learning framework TL-SSGAN that integrates TL and GAN, is proposed here, as shown in Fig.11. The TL models can be pre-trained or fine-tuned. And the ratio of labeled and unlabeled samples can be used as a variable parameter, additional unlabeled samples may also be added. The output of discriminator/classifier (D/C) of TL-SSGAN will be whether the test samples are generated or real and whether it is a crack sample or not.

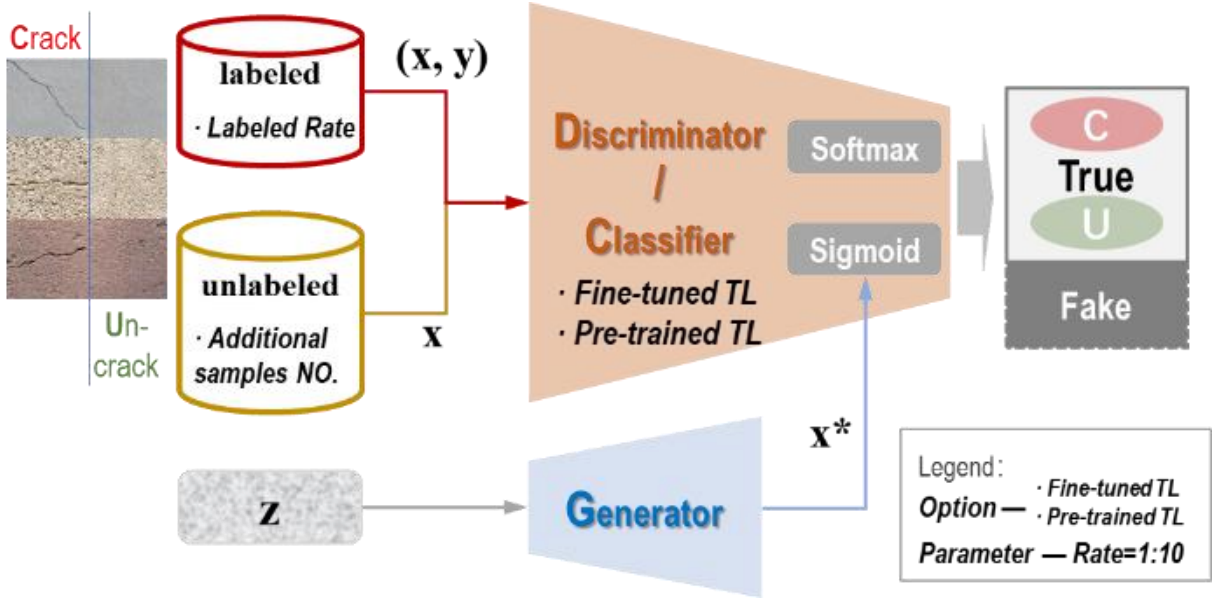

Fig. 11. Framework and pipeline of Semi-supervised GAN embedded TL models as encoder (D/C).

Table 5. Ablation test on performance of the Proposed TL-SSGAN on SDNET2018.

| TL-SSGAN Framework with Different Models and Data | | | Accuracy | Precision | Recall | F1-score |
|---|---|---|---|---|---|---|
| Column NO. | TL Model (Backbone, fine-tuned or not) | Labeled Rate (Labeled/Total) + Additional Samples | | | | |
| 1 | Resnet152V2, pre-trained | 1:30 | 0.888 6 | 0.926 7 | 0.945 3 | 0.934 9 |
| 2 | Resnet152V2, pre-trained | 1:20 | 0.889 2 | 0.931 4 | 0.947 8 | 0.939 5 |
| 3 | Resnet152V2, pre-trained | 1:10 | 0.905 9 | 0.939 6 | 0.955 3 | 0.946 5 |
| 4 | Resnet152V2, fine-tuned | 1:30 | 0.901 6 | 0.929 3 | 0.956 7 | 0.942 8 |
| 5 | Resnet152V2, fine-tuned | 1:20 | 0.906 5 | 0.930 0 | 0.958 9 | 0.944 2 |
| 6 | Resnet152V2, fine-tuned | 1:10 | 0.924 5 | 0.941 9 | 0.974 9 | 0.956 7 |
| 7 | InceptionV3, fine-tuned | 1:10 | 0.917 0 | 0.945 5 | 0.961 0 | 0.953 2 |
| 8 | MobileNetV2, fine-tuned | 1:10 | 0.910 6 | 0.936 2 | 0.959 0 | 0.945 6 |
| 9 | Resnet152V2, fine-tuned | 1:10 + Additional 10k unlabeled samples | 0.931 4 | 0.946 9 | 0.972 1 | 0.959 3 |
| 10 | Resnet152V2, fine-tuned | 1:10 + Additional 20k unlabeled samples | 0.938 8 | 0.957 0 | 0.972 3 | 0.964 6 |

Evaluation tests were performed on SDNET2018 and the results of TL-SSGAN algorithms are shown in Table 5. It can be observed form Table 5 that:

(1) Models based on TL-SSGAN framework (Fig.11) have achieved good patch classification performance on SDNET2018. In our test, the best accuracy, precision, recall and F1-score are respectively 0.938 8, 0.957 0, 0.972 3, 0.964 6, which are close to the results of the supervised TL (seen in Table 3).

(2) Both backbone model and fine-tuned mechanism in the TL framework contribute significantly to the improvement of accuracy. For example, all other factors or parameters being same, Resnet152V2-based algorithm is 1.5%, 0.6%, 1.7%, and 1.2% better than MobileNetV2-based respectively on metrics of accuracy, precision, recall, and $F_1$-score (i.e., Table5 Column6 vs. 8); while the results of fine-tuned algorithm increase about 1.5~2.1%, -0.2~0.3%, 1.2~2.1%, 0.5~1.1% respectively, compared to pre-trained algorithm, when both with Resnet152V2 as backbone (i.e., Table5 Column4-6 vs. Column1-3 respectively).

(3) In the aspect of data usage, when the ratio of labeled/unlabeled sample number is 1:10, compared to 1:30 (Resnet152V2 as backbone), the metrics' results increase respectively about 1.9~2.5%, 1.4%, 1.1~1.9%, 1.2~1.5% (i.e., Table5 Column3 vs. 1 and Column6 vs. 4). By adding additional unlabeled samples up to 2 times, an increase of about 0.3~1.6% can further be achieved (i.e., Table5 Column9,10 vs. 6). Therefore, the weakly supervised mechanism of proposed TL-SSGAN framework can not only reduce the dependence on labeled samples, but also improve the classification performance by using incremental unlabeled samples.

## 4. Summary and Conclusion

In general, the patch classification performance of the fine-tuned TL algorithms on crack public data sets (e.g., CFD, SDNET2018), is comparable to or slightly better than that of ED algorithms, in terms of accuracy (precision, precision, recall, and $F_1$-score), and predicting time cost is less (about 8.1~50.2ms/patch). It is because TL algorithms benefit from the architecture-optimized backbone models pre-trained on large scale data sets. These backbones can also be used as the basic network in ED and GAN architectures. In terms of accurate crack location, ED (e.g., FPCNet) and GAN (e.g., CrackGAN) can achieve pixel-level segmentation. In terms of detection efficiency, CrackGAN only takes about 1.6s to predict a 4096x2048 full-size image on NVIDIA 1080TI GPU. The TL method based on lightweight backbone (e.g., MobileNetV2) is even faster. Thus, it is expected to realize real-time detection of crack identification (classification/segmentation) on low-computing platform.

In real practice, due to 1) Subjective factors: inaccuracy of manual annotation, time-consuming and laborious large-scale sample labeling; and 2) Objective factors: there are lots of topological complex cracks and thin cracks in the image samples collected in industrial applications (e.g., pavement crack images collected by high-speed running vehicle, and defects images of industrial products on automatic assembly line), which will lead to bias in the evaluation by using quantified indicators such as accuracy, precision, recall, $F_1$-score. Thus, HD-score is advised as quantitative evaluation indicator for crack pixel-level identification, which owns good discrimination when crack ground truth (GT) semi-accurate labeling (e.g., 1-pixel curve labeling).

GT semi-accurate labeling is a labor-light efficient ways to reduce the workload of manual annotation. We proposed a weakly supervised learning framework combining TL and semi-supervised GAN, through the measures of utilizing fine-tuned TL backbone model, controlling the ratio of labeled samples and unlabeled ones, and adding additional unlabeled samples, which can maintain comparable crack identification performance with the supervised learning while greatly reducing the number of labeled samples used.

To sum up, the combination of various deep learning frameworks such as TL, ED and GAN, can integrate the advantages of each respective mechanism and improve the performance of the overall architecture.

## Acknowledgments

The author would like to thank Dr. Ni Wei and Ni Yong for their support, discussion and inspiration.

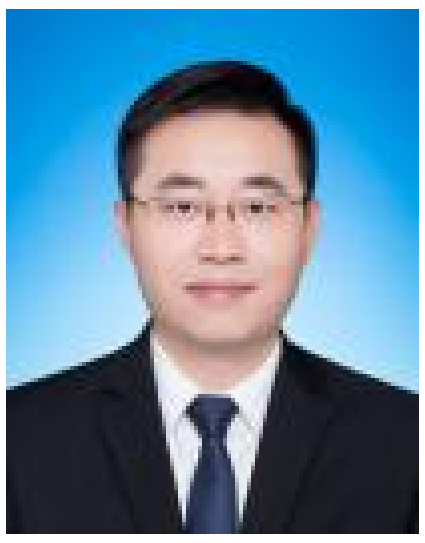

**Kai-Liang LU** received his Ph.D. from Tongji University in 2010. From 2011 to 2018, he was a Lecturer and later Associate Professor since 2013 at Shanghai Maritime University. From 2018, he worked for Huawei Technologies Co. Ltd etc. as a Senior Engineer. Now, he is an Algorithm Researcher at Jiangsu Automation Research Institute Shanghai Branch.

Kai-Liang LU is the author of over 20 technical publications, including journal articles, proceedings, and standards. His research interests include engineering vibration FEM analysis, computer vision and pattern recognition, computer aided intelligence. He serves as a Committee Member of Vibration Mechanics for Shanghai Society of Theoretical and Applied Mechanics. He is an active member of China Computer Federation (CCF).